\documentclass[10pt,letterpaper]{article}
\usepackage[top=0.85in,left=2.75in,footskip=0.75in,marginparwidth=2in]{geometry}

\usepackage[T1]{fontenc}

\usepackage{cite}
\usepackage{covington}
\usepackage{nameref,hyperref}

\usepackage{microtype}
\DisableLigatures[f]{encoding = *, family = * }

\raggedright
\usepackage{changepage}

\usepackage[aboveskip=1pt,labelfont=bf,labelsep=period,singlelinecheck=off]{caption}

\makeatletter
\renewcommand{\@biblabel}[1]{\quad#1.}
\makeatother

\usepackage{lastpage,fancyhdr,graphicx}
\usepackage{epstopdf}
\fancyheadoffset[L]{2.25in}
\fancyfootoffset[L]{2.25in}

\usepackage{color}

\definecolor{Gray}{gray}{.25}

\usepackage{graphicx}

\usepackage{sidecap}

\usepackage{wrapfig}
\usepackage[pscoord]{eso-pic}
\usepackage[fulladjust]{marginnote}
\reversemarginpar

\begin{document}
\vspace*{0.35in}

% title goes here:
\begin{flushleft}
{\Large
\textbf\newline{Exploiting Cross-Dialectal Gold Syntax for Low-Resource Historical Languages: Towards a Generic Parser for Pre-Modern Slavic.}
}
\newline
% authors go here:
\\
Nilo Pedrazzini\footnote{Preprint: to appear in \textit{Proceedings of the Workshop on Computational Humanities Research} (CHR 2020).}
\\
\bigskip
\bf University of Oxford
\\
\bigskip
\verb|nilo.pedrazzini@ling-phil.ox.ac.uk|

\end{flushleft}

\section*{Abstract}
This paper explores the possibility of improving the performance of specialized parsers for pre-modern Slavic by training them on data from different related varieties. Because of their linguistic heterogeneity, pre-modern Slavic varieties are treated as low-resource historical languages, whereby cross-dialectal treebank data may be exploited to overcome data scarcity and attempt the training of a variety-agnostic parser. Previous experiments on early Slavic dependency parsing are discussed, particularly with regard to their ability to tackle different orthographic, regional and stylistic features. A generic pre-modern Slavic parser and two specialized parsers --one for East Slavic and one for South Slavic-- are trained using jPTDP \cite{nguyen}, a neural network model for joint part-of-speech (POS) tagging and dependency parsing which had shown promising results on a number of Universal Dependency (UD) treebanks, including Old Church Slavonic (OCS). With these experiments, a new state of the art is obtained for both OCS (83.79\% unlabelled attachment score (UAS) and 78.43\% labelled attachement score (LAS)) and Old East Slavic (OES) (85.7\% UAS and 80.16\% LAS).

% the * after section prevents numbering

\section{Parsing data-poor historical languages: the case of Slavic}
Low-resource languages represent a considerable challenge in Natural Language Processing (NLP), which is notoriously data-demanding. Data-poor languages and Big Data thus generally call for very different methodologies. Some languages may be ‘low-resource' because they have only recently been recorded for the first time, with the ensuing difficulty of dealing with an ad hoc writing system or no standard orthography altogether, as is the case for many currently endangered languages \cite{essegbey}. Others may be widely spoken or relatively well-documented, but hardly have a sufficient amount of structured data (e.g. large, manually labelled corpora) to be target languages in downstream NLP tasks (e.g. \cite{ojha}). 
\\ \indent Low-resource historical languages present additional hurdles: they are closed sets and they necessarily lack native-speaker inputs. Unlike low-resource languages which can rely on data collection by means of fieldwork or on the expansion of manually annotated corpora by native speakers, the creation of structured data for historical languages is dependent on the digitisation of written sources that are virtually never only found in a contained geographic area. Even when digital editions are available, historical languages often lack a unified literary standard, which may result in linguistic and orthographic inconsistencies across and within individual texts. High orthographic variation, for instance, is obviously not ideal for the training of NLP models of a language which is already under-resourced to begin with. 
\\ \indent Pre-modern Slavic varieties\footnote{‘Pre-modern Slavic' here assumes early Slavic varieties of the so-called Slavia Orthodoxa \cite{picchio}, that is, chiefly excluding all West Slavic languages (the Slavic subgroup which includes contemporary Czech, Slovak and Polish).} are illustrative in this respect. Two dialect macro-areas, East and South Slavic, can be distinguished ever since the earliest Slavic sources (10-11th century) on the basis of various phonological and morphological features. However, the subvarieties belonging to each group have often very distinct characteristics. In an ideal world, the development of powerful tools for the processing of each dialect area would be carried out by using an equally large amount of data from all varieties. Table \ref{tab1} shows which pre-modern Slavic varieties in the TOROT Treebank \cite{eckhoff2015a} \cite{eckhoff2018a} contain morphological and dependency annotation that could potentially be exploited for the development of NLP tools\footnote{A detailed breakdown of all the texts in corpus (including the labels with which they are referred to throughout the paper), with an indication on their language variety and number of tokens, can be found in the Appendix.}. Not only is there a disproportion between the two dialect areas (with East Slavic being preponderant), but their subvarieties are far from being evenly distributed. Some major early Slavic varieties are not represented at all (e.g. Middle Bulgarian), which is also due to the fact that manual annotation can be slower or faster depending on the amount of secondary sources that may help speed up the process (e.g. translations and critical editions).

\begin{table*}[ht]
\centering
    \caption{Early Slavic dialect macro-areas and subvarieties with gold syntactic annotation in the TOROT Treebank}
    \label{tab1}
    \begin{tabular}{cccccl}
    \hline
Dialect macro-areas  &   Varieties  & Label & Tokens  \\
    \hline
South Slavic&  Old Church Slavonic   &  OCS & 139,055 \\
 &  Serbian Church Slavonic   &  SCS & 890\\
 &  Russian Church Slavonic   &  RCS  & 331 \\
     \hline
 East Slavic &  Old East Slavic   &  OES & 142,138\\
 &  Middle Russian   &  MRus  & 95,066\\
 &  Old Novgorodian  &  ONov  & 2,245\\
    \hline
  \end{tabular}
\end{table*}

Computational techniques for the processing of early Slavic sources have been developing relatively quickly, including tools tackling the issue of obtaining digital primary sources (e.g. neural networks for handwritten text recognition \cite{rabus2019}). The state of the art in automatic POS and morphological tagging has also reached success rates nearly as high as those of contemporary-language taggers \cite{scherrer2018a}. By contrast, syntactic annotation is still performed almost exclusively manually. The result is that several texts in the corpus contain either no gold dependency annotation, or morphological tagging only. This arguably defies the very purpose of digital corpora of low-resource languages, which may be expected to contain detailed annotation throughout, precisely because they are necessarily limited in size. Besides, the implementation of annotation schemes pertaining to linguistic levels deeper than syntax (e.g. information and discourse structure) fully relies on having high-quality syntactic annotation in the first place. Even more importantly, syntactically annotated corpora can be exploited for corpus-driven typological analyses, which can be crucial to advance linguistic theory. The disparity between low- and high-resource languages with regard to the availability of such resources thus risks to generate a bias towards patterns observed in the latter.

\subsection{State of the art in pre-modern Slavic dependency parsing}
Previous attempts at developing parsers for pre-modern Slavic have only been carried out on one of its dialect areas or on specific subvarieties:
\begin{itemize}
    \item In \cite{eckhoff2016}, a parser for OES was trained using MaltParser \cite{nivre} and was shown to be an efficient pre-annotation tool, yielding a decent annotation speed gain, but with a considerable difference between experienced and inexperienced annotators. However, as the authors note, its best scores (84.5\% UAS and 73.4\% LAS) were likely due to the simple genre and to the few long sentences of the test set. To the best of my knowledge, the experiment still represents the state of the art in the automatic parsing of OES.
    \item An off-the-shelf parser for OCS is instead available from UD \cite{nivre}\footnote{\url{http://ufal.mff.cuni.cz/udpipe/models}}. The model, which reached relatively high scores (80.6\% UAS and 73.4\% LAS) was however only trained and tested on a single text (viz. \textsc{marianus}). As a result, these scores do not reflect real-world performance\footnote{In this context, ‘real-world performance' refers to how well a model deals with texts that present different orthographic, regional and stylistic features} and the parser is hardly applicable to texts falling outside the set of orthographic and linguistic peculiarities of \textsc{marianus}, which are only shared to some extent by the other texts classified as ‘OCS' in Table \ref{tab1}.
    \item Finally, a neural network model has recently been trained on a number of UD treebanks, including OCS, using bidirectional long-short memory (BiLSTM) to jointly learn POS tagging and dependency parsing \cite{nguyen} (jPTDP). Its results for dependency parsing are similar to those of the off-the-shelf UD baseline OCS parser, but with a slight LAS improvement (+0.5\%), thus representing the state of the art for OCS. Nevertheless, the same issue pointed out about the UD baseline parser applies: the scores given in \cite{nguyen} only refer to \textsc{marianus}, which makes the model unusable beyond OCS texts that present orthographic and linguistic features very close to those of \textsc{marianus} itself.
\end{itemize}

\subsection{Aims of this paper}
The goal of this paper is twofold: 
\begin{itemize}
    \item To investigate the extent to which the performance of specialized (i.e. variety-specific) parsers can be improved by expanding the training set with data from other varieties and dialect areas.
    \item To explore the possibility of attaining a ‘generic’ parser, a tool which is relatively dialect-agnostic and more flexible with respect to genres and historical stages.
\end{itemize}

A generic parser could especially speed up the annotation of pre-modern Slavic texts whose language and orthography are not straightforwardly classifiable in terms of provenance. Early Slavic texts written in hybrid varieties are in fact rather the norm than the exception, which is due to intricate manuscript traditions, to the lack of a unified written standard, and to the complex relationship between vernacular(s) and literary language(s). 
\\ \indent This experiment attempts to enhance the real-world performance of jPTDP \cite{nguyen}, by training it on three different datasets: one containing only South Slavic data (OCS, RCS and SCS), one only East Slavic data (OES, MRus and ONov), and one both macro-varieties. The choice of retraining jPTDP rather than attempting to develop a novel neural network model is motivated, on the one hand, by the fact that jPTDP seems to perform particularly well with morphology-rich languages, which is the case for Slavic; on the other hand, we are interested in noting how the addition of heterogeneous training data affects its performance on OCS, which is the only pre-modern Slavic variety represented among the UD treebanks. Besides, jPTDP has not been tested on early East Slavic data, which allows us to compare the performance of a neural network model to that of MaltParser.
\\ \indent Section \ref{sec2} outlines the pre-processing stage, including the criteria used to split the corpus into training, development and test sets. Section \ref{sec3} lays out the training of jPTDP, including the choice of hyperparameters, and compares the results obtained for the three parsers during cross-validation. Section \ref{sec4} is dedicated to the evaluation of the parsers by means of test sets which are meant to be indicative of real-world performance. Conclusions then follow with suggestions for future experiments. 

\section{Pre-processing} \label{sec2}

All the data used in this experiment comes from the latest TOROT Treebank release\footnote{\url{https://github.com/torottreebank/treebank-releases/releases/tag/20200116}}. The corpus includes pre-modern Slavic text spanning from the oldest Slavic attestations (10th-11th century) to OES and MRus texts from the 11th-19th century \cite{eckhoffberdi2}. It also includes a contemporary Russian subcorpus, which was however left out since we are only interested in the early stages of Slavic.
\\ \indent In order to reach representativeness and limit overfitting, 10\% of each text was set aside as development data (40,375 tokens), 10\% as test data (39,886 tokens) and 80\% as training data (240,571 tokens). Texts with fewer than 400 tokens were exclusively employed for training\footnote{This number (i.e. 400 tokens) was simply the minimum which allowed to split each text with a 80:10:10 proportion.}. By doing so, we obtained a relatively homogeneous distribution of genres and language varieties. Only for \textsc{marianus} the predefined UD split into training, development and test set was adopted, to allow a comparison between our results and those of \cite{nguyen}.
\\ \indent The training, development and test portions of each text are kept separate and merged only at need, which allows for faster experimentation with different combinations of texts while keeping the proportions consistent throughout\footnote{All the datasets used in this experiment can be found at \url{https://doi.org/10.6084/m9.figshare.12950093.v1}. These include separate training, development and test files for each individual text.}.
\\ \indent TOROT releases come in two formats: the standard PROIEL XML format and the CoNLL-X format of UD. jPTDP requires the updated CoNLL-U format as input, whose main differences with CoNLL-X are the treatment of multiword tokens as integer ranges and the insertion of comments before each new sentence, besides the different order and outlook of their morphotags (e.g. NUMBs|GENDn|CASEn in CoNLL-X and Case=Nom|Gender=Neut|Number=Sing in CoNLL-U). The datasets were converted from PROIEL XML to CoNLL-U using the script included in the Ruby utility \verb|proiel-cli|, which can be used for the manipulation of PROIEL treebanks\footnote{\url{https://github.com/proiel/proiel-cli}}.

\section{Training}\label{sec3}

In the first round of training, jPTDP was applied directly off-the-shelf with its default hyperparameters, in order to compare the scores in \cite{nguyen} with those resulting from our larger training set: 30 training epochs, 50-dimensional character embeddings, 100-dimensional word embeddings, 100-dimensional POS tag embeddings, 2 BiLSTM layers, 128-dimensional LSTM hidden states and 100 hidden nodes in each one-hidden-layer multi-layer perceptron (MLP). The hyperparameters were thus set by the authors of \cite{nguyen} on the basis of the optimal hyperparameters for the English WSJ Penn Treebank \cite{marcus}, which they established through a minimal grid search and applied to all UD treebanks without individual optimization. The only exception is the default size of LSTM hidden states, which they fixed at 128, even though the optimal value on the English WSJ Penn Treebank was found to be 256. 
\\ \indent In the second round of training a grid search was performed to select the optimal size of LSTM hidden states in each layer from \{128, 256\} and the number of hidden nodes in MLPs from \{100, 200, 300\}. Due to limited computational resources, the other hyperparameters were set to default.
\\ \indent While the experiment in \cite{nguyen} suggests a better performance for jPDTP using 256-dimensional LSTM hidden states, our results during cross-validation indicate that this is not necessarily the case with pre-modern Slavic data. As Table \ref{tab2} shows, only the generic model (jPTDP-GEN) benefits from a larger number of BiLSTM dimensions, whereas the specialized models, both the South Slavic (jPTDP-SSL) and the East Slavic one (jPTDP-ESL), perform better with a larger number of hidden nodes in MLPs (300), but 128 BiLSTM dimensions.

\begin{table}[ht]
\centering
\caption{Highest scores obtained during cross-validation using the optimal hyperparameters for each dataset}
\label{tab2}
\begin{tabular}{cccccc}
\hline
Model  & LSTM hidden states size & MLP hidden layer size & LAS  & UAS  \\
\hline
jPTDP-SSL & 128 & 300  & 71.10 & 78.95 \\
jPTDP-ESL  & 128 & 300  & 73.65 & 79.95 \\
jPTDP-GEN  & 256 & 200  & 72.07 & 79.39 \\
\hline
\end{tabular}
\end{table}

In Section \ref{sec4} separate evaluations of the models developed with default and optimized hyperparameters will be provided for the sake of comparison. The evaluation phase will also show not only that real-world performance varies greatly depending on the text, but also that the scores emerged during cross-validation do not reflect the relative quality of the trained parsers. In all likelihood, this is primarily due to the fact that the development sets are virtually fully homogeneous, linguistically and stylistically, with the respective training sets, since they both comprise a percentage of nearly all texts written in the relevant Slavic variety.

\section{Evaluation} \label{sec4}
\begin{table}[ht]
\centering
\caption{Test sets description}
\label{tab3}
\begin{tabular}{cccc}
\hline
Label & Dialect macro-area & Varieties & Texts \\
\hline
\textsc{ss} & South Slavic & OCS, SCS & All south Slavic texts ($>$400 tokens)\\
\textsc{cm} & South Slavic & OCS & \textsc{marianus} \\ 
\textsc{cs} & South Slavic & OCS & \textsc{supr} \\
\textsc{vc} & South Slavic & SCS & \textsc{vit-const} \\
\textsc{es} & East Slavic & OES, MRus, ONov & All east Slavic texts ($>$400 tokens) \\
\textsc{pc} & East Slavic & OES & \textsc{lav}  \\
\textsc{sr} & East Slavic & MRus & \textsc{sergrad}  \\
\textsc{av} & East Slavic & MRus & \textsc{avv}  \\
\textsc{on} & East Slavic & ONov & \textsc{birchbark}  \\
\hline
\end{tabular}
\end{table}

Each parser was tested on nine datasets which were chosen as representative of distinct early Slavic varieties and historical stages (Table \ref{tab3}). In particular:

\begin{itemize}
    \item \textsc{ss} and \textsc{es}, containing 10\% of all East and South Slavic text respectively, are meant to show the performance of the parsers on the relevant dialect macro-areas as a whole.
    \item \textsc{cm} corresponds to the test set of both the UD baseline OCS parser and \cite{nguyen}.
    \item \textsc{cs} is used to compare the performance of the parsers on OCS texts other than \textsc{marianus}. As a miscellany, the syntax of \textsc{supr} is more varied than \textsc{marianus}, which exclusively contains OCS translations of the Gospels. Moreover, though both very archaic (i.e. relatively close to reconstructed Proto-Slavic), they present different regional features (Bulgarian in \textsc{supr}, Macedonian in \textsc{marianus}) and reflect different manuscript traditions (\textsc{marianus} is a Glagolitic manuscript, \textsc{supr} a Cyrillic one).
    \item \textsc{vc} is used as the only late South Slavic manuscript (16th century) with clear Serbian features.
    \item \textsc{pc} is one of the most important OES manuscripts and the test sets used by \cite{eckhoffberdi}.
    \item \textsc{sr} and \textsc{av} represent two distinct varieties of MRus. The language of the former is in fact often classified as RCS, because of its hybrid Church Slavonic and Russian features. The latter is instead a 17th-century Russian text heavily influenced by the vernacular language.
    \item \textsc{on} is representative of ONov, which is not only notoriously distinct from the Old Kievan and Moscovite varieties of early east Slavic (OES/MRus), but it also mostly consists of vernacular material -- as opposed to the remaining East Slavic texts in the corpus, often heavily influenced by Church Slavonic (i.e. South Slavic). 
\end{itemize}  

The evaluation script which was used to compare gold and predicted tags can be found in the official UD repository\footnote{\url{https://universaldependencies.org/conll18/evaluation.html}}. 
\\ \indent As Table \ref{tab4} shows, jPTDP-GEN performs best on all South Slavic test sets (\textsc{ss}, \textsc{cm}, \textsc{cs}, \textsc{vc}), as well as on the East Slavic \textsc{av} and \textsc{sr} datasets. However, even when jPTDP-ESL performs better (viz. on \textsc{es}, \textsc{pc} and \textsc{on}) jPTDP-GEN does not lag far behind. This clearly indicates that cross-dialectal training data may improve the performance of a parser, even if it is meant to be used to annotate only text of a particular variety. 

\begin{table}[ht]
\centering
\caption{Models evaluation: UAS-d[efault] and LAS-d[efault] are the scores obtained from the models trained with default hyperparameters, whereas UAS and LAS are those obtained from the optimized models, as defined in Table \ref{tab2}.}
\label{tab4}
\begin{tabular}{cccc}
\hline
Test Set & Model & UAS & LAS \\
\hline
\textsc{ss} & jPTDP-SSL & 76.99 & 69.51 \\
 & jPTDP-ESL & 72.94 & 62.61 \\
 & jPTDP-GEN & \textbf{78.86} & \textbf{71.87} \\ \hline
\textsc{cm} & jPTDP-SSL & 83.61 & 77.98 \\
 & jPTDP-ESL & 83.60 & 77.83 \\
 & jPTDP-GEN & \textbf{83.79} & \textbf{78.42} \\ \hline
\textsc{cs} & jPTDP-SSL & 68.54 & 58.76 \\
& jPTDP-ESL & 58.92 & 42.88 \\
 & jPTDP-GEN & \textbf{72.28} & \textbf{63.38} \\ \hline
\textsc{vc} & jPTDP-SSL & 61.54 & 51.28 \\
 & jPTDP-ESL & 66.67 & 48.72 \\
 & jPTDP-GEN & \textbf{69.23} & \textbf{56.41} \\ \hline
\textsc{es} &  jPTDP-SSL & 62.83 & 47.55 \\
 & jPTDP-ESL & \textbf{81.02} & \textbf{74.93} \\
 & jPTDP-GEN & 80.86 & 74.23 \\  \hline
\textsc{pc} & jPTDP-SSL & 68.08 & 52.08 \\
 & jPTDP-ESL & \textbf{85.70} & \textbf{80.16}  \\
 & jPTDP-GEN & 85.22 & 79.29 \\ \hline
\textsc{sr} & jPTDP-SSL & 58.63 & 41.42 \\
 & jPTDP-ESL & 71.59 & 63.91 \\
 & jPTDP-GEN & 73.24 &  64.71 \\ \hline
\textsc{av} & jPTDP-SSL & 62.08 & 45.25 \\
 & jPTDP-ESL & 80.91 &  74.96 \\
 & jPTDP-GEN & \textbf{81.75} & \textbf{75.80} \\\hline
\textsc{on} & jPTDP-SSL & 58.82 & 41.18 \\
& jPTDP-ESL & \textbf{74.33} & \textbf{58.82} \\
& jPTDP-GEN & 72.19 & 58.29 \\
\hline
\end{tabular}
\quad
\begin{tabular}{cc}
\hline
UAS-d & LAS-d \\
\hline
 77.08 & 69.54  \\ 
 61.29 & 45.11  \\ 
 78.11 & 70.72  \\ \hline
 83.19 & 77.63  \\ 
 66.03 & 50.98  \\ 
 83.32 & 77.79  \\ \hline
 69.21 & 59.30  \\ 
 54.62 & 37.13  \\
 71.22 & 61.53  \\ \hline
 60.26 & 51.28  \\ 
 60.26 & 43.59  \\ 
 61.54 & 50.00  \\ \hline
 63.18 & 47.51  \\ 
 80.74 & 74.44  \\ 
 80.67 & 74.18  \\ \hline
 66.52 & 50.86  \\ 
 85.59 & 79.25  \\
 84.51 & 78.69  \\ \hline
 58.84 & 41.16  \\ 
 71.34 & 63.50  \\ 
 \textbf{73.90} & \textbf{65.76}  \\ \hline
 61.01 & 44.45  \\ 
 79.44 & 73.89  \\ 
 81.22 & 75.44  \\ \hline
 57.75 & 41.18  \\ 
 71.66 & 55.61  \\
 68.98 & 53.48  \\ 
\hline
\end{tabular}
\end{table}

Unsurprisingly, there are also obvious indications that the level of representativeness of a variety among the training data has important consequences on the performance of the parsers. The scores obtained on \textsc{vc} and \textsc{on} are particularly low, with LAS $<$ 60.00. This is likely due to the fact that SCS and ONov linguistic features are greatly underrepresented in the corpus. Expanding the training data with these varieties is therefore very likely to improve the performance of the models. 
\\ \indent Several errors are due to orthographic idiosyncrasies of the individual manuscripts (or the edition thereof), whereby an unusual spelling may render the syntactic relation of a word ambiguous. In (\ref{ex1}), for instance, the word \textit{ōtinoud} ‘not at all' is spelt differently from its most usual forms, \textit{otinoudĭ} or \textit{ōtinudĭ}. It is likely that the parser expected a singular subject in its position, given the following main verb \textit{vozdrŭžaše} ‘(he) abstained'. The lack of final -\textit{ĭ} in \textit{ōtinoud}  does in fact make the word appear morphologically like a singular masculine noun.

\begin{covexample}
\glll i ōt pijęnĭstva ōtinoud vozdrŭžaše sę
and.\textsc{cc} from.\textsc{case} drinking.\textsc{obl} not-at-all.\textsc{advmod} abstain.\textsc{root} himself.\textsc{aux} (Gold)
and.\textsc{cc} from.\textsc{case} drinking.\textsc{obl} not-at-all.\textsc{nsubj} abstain.\textsc{root} himself.\textsc{aux} (Predicted)
\glt ‘And by no means did he abstain from drinking' (\textsc{sergrad}, 17r)
\glend
\label{ex1}
\end{covexample}

It is particularly interesting to note that the scores obtained on \textsc{cs}, the only other OCS text in the corpus, are not as high as those reached with \textsc{cm}, which corresponds to the test set in \cite{nguyen}. This is indicative of the fact that the previous state of the art in parsing OCS does not reflect real-world performance. In our case, the relatively low scores obtained on \textsc{cs} appear to be mostly due to its more complex syntactic structures compared to \textsc{cm}. In (\ref{ex2}), for instance, the indirect object is repeated twice, once after the subject and once just before the main verb, which is plausibly the main cause of the poor performance of the parser on the rest of the sentence:

\begin{covexample}
\glll i jeliko tebě ljubo i drago tebě bǫde 
and.\textsc{cc} whoever.\textsc{nsubj} to-you.\textsc{iobj} beloved.\textsc{nsubj} and.\textsc{cc} dear.\textsc{conj} to-you.\textsc{iobj} will-be.\textsc{root} (Gold)
and.\textsc{cc} whoever.\textsc{advmod} to-you.\textsc{advmod} beloved.\textsc{xcomp} and.\textsc{advmod} dear.\textsc{xcomp} to-you.\textsc{obl} will-be.\textsc{root} (Predicted)
\glt ‘And whoever will be beloved and dear to you' (\textsc{supr}, 24v)
\glend
\label{ex2}
\end{covexample}

As Tables \ref{tab5} and \ref{tab6} show, with our models we obtained a new state of the art in both OCS and OES dependency parsing. It is worth noting that while jPTDP-ESL performed slightly better on \textsc{pc}, jPTDP-GEN is also past the state of the art for OES. This is a particularly promising result: as already discussed, because of the lack of standardization in pre-modern Slavic, the development of a high-quality generic parser should arguably be prioritized over that of multiple specialized models. While this could mean a slightly lower performance than specialized parsers when it comes to well-represented varieties (e.g. OES), the long-term benefit of a dialect-agnostic tool are likely to be more substantial. A decent-quality generic parser could in fact be employed to speed up the annotation of underrepresented varieties, which would ultimately result in the expansion of deeply annotated treebanks.

\begin{table}[ht]
\centering
\caption{OCS: comparison with previous experiments on \textsc{marianus}}
\label{tab5}
\begin{tabular}{cccc}
\hline
Model & UAS & LAS \\
\hline
UD baseline model  & 80.6 & 73.4 \\
jPTDP \cite{nguyen}  & 80.59 & 73.93 \\
jPTDP-GEN (default hyperparameters)  & 83.32 & 77.79 \\
jPTDP-GEN (optimized) & \textbf{83.79} & \textbf{78.42} \\
\hline
\end{tabular}
\end{table}

\begin{table}[ht]
\centering
\caption{OES: comparison with previous experiments on \textsc{lav}}
\label{tab6}
\begin{tabular}{cccc}
\hline
Model  & UAS & LAS \\
\hline
Maltparser \cite{eckhoffberdi}  & 84.5 & 77.9 \\
jPTDP-ESL (default hyperparameters)  & 85.59 & 79.25 \\
jPTDP-ESL (optimized) & \textbf{85.7} & \textbf{80.16} \\
\hline
\end{tabular}
\end{table}

\section{Conclusions and future experiments}
This paper explored the possibility of exploiting syntactically annotated treebanks of related but distinct pre-modern Slavic varieties in order to train a generic, variety-agnostic parser. The results suggest that the performance of a specialized model can in fact be considerably improved by expanding the training data with different pre-modern Slavic varieties. This has particularly emerged from the scores obtained on OCS (South Slavic) by the generic parser, which was trained on both South and East Slavic data. With our experiment a new state of the art has been obtained for both OCS (UAS 83.79\% and LAS 78.43\%) and OES (UAS 85.7\% and LAS 80.16\%).
Future studies may wish to attempt larger-scale experimentation with cross-lingual transfer learning across different related historical languages. The automatic processing of OCS is especially very likely to benefit from direct transfer or annotation projection, as well as from cross-lingual word representations, from Ancient and New Testament Greek, given the comparatively very similar linguistic systems of Slavic and Greek. 

\section*{Acknowledgments}
This work was supported by the Economic and Social Research Council [grant number ES/P000649/1]. I am grateful to Marius Jøhndal for his help with Ruby while setting up the \verb|proiel-cli| utility.
%This is where your bibliography is generated. Make sure that your .bib file is actually called library.bib

%This defines the bibliographies style. Search online for a list of available styles.
\bibliographystyle{abbrv}
%%
%% Define the bibliography file to be used

%%
%% If your work has an appendix, this is the place to put it.
\appendix
\section{Appendix}
Table \ref{tab:appen} below contains a detailed breakdown of the datasets used in the experiment.
\begin{table}
\caption{Dataset breakdown, with an indication of the language variety represented by each manuscript. The text labels reproduce the codes used by the official TOROT releases, to facilitate text retrieval should one wish to check the results of this paper against the original datasets.}
  \label{tab:appen}
    \begin{tabular}{cccc}
    \hline
 Variety &  Text &  Label & Tokens  \\
    \hline
    OCS  & Codex Marianus  & \textsc{marianus} & 58,269  \\
  & Codex Suprasliensis  & \textsc{supr} & 79,070  \\
  & Codex Zographensis  & \textsc{zogr} & 1,098  \\
  & Kiev Missal  & \textsc{kiev-mis} & 370  \\
  & Psalterium Sinaiticum  & \textsc{psal-sin} & 248  \\
SCS  & Vita Constantini  & \textsc{vit-const} & 890  \\
RCS  & Vita Methodii  & \textsc{vit-meth} & 331  \\
    OES  & Primary Chronicle (Codex Laurentianus)  & \textsc{lav} & 56,725  \\
  & Suzdal Chronicle (Codex Laurentianus)   & \textsc{suz-lav} & 23,760  \\
  & Primary Chronicle (Codex Hypathianus)   & \textsc{pvl-hyp} & 3,610  \\
   & First Novgorod Chronicle (Synodal)   & \textsc{nov-sin} & 17,838  \\
  & Kiev Chronicle (Codex Hypathianus)   & \textsc{kiev-hyp} & 544  \\
  & Colophon (Mstislav's Gospel)  & \textsc{mstislav-col} & 259  \\
  & Colophon (Ostromir Codex)  & \textsc{ostromir-col} & 199  \\
  & Missive (Archbishop of Riga)   & \textsc{rig-smol1281} & 171  \\
  & Mstislav's letter   & \textsc{mst} & 158  \\
  & Novgorod’s treaty with Jaroslav   & \textsc{novgorod-jaroslav} & 423  \\
  & Russkaja pravda   & \textsc{rusprav} & 4,174  \\
  & Statute of Prince Vladimir   & \textsc{ust-vlad} & 495  \\
  & Treaty (Smolensk-Riga-Gotland)   & \textsc{riga-goth} & 1,421  \\
  & The Tale of Igor’s Campaign   & \textsc{spi} & 2,850  \\
  & Russkaja pravda   & \textsc{rusprav} & 4,174  \\
  & Uspenskij Sbornik (excerpts)  & \textsc{usp-sbor} & 25,189  \\
   & Varlaam Xutynskij's Grant Charter & \textsc{varlaam} & 148  \\
MRus  & Afanasij Nikitin's \textit{Journey}  & \textsc{afnik} & 6,842  \\
& Charter of Prince Jurij Svjatoslavich   & \textsc{smol-pol-lit} & 344  \\
& Correspondence of Peter the Great   & \textsc{peter} & 100  \\
& Domostroj   & \textsc{domo} & 23,459  \\
& Life of Sergij of Radonezh   & \textsc{sergrad} & 20,361  \\
& History of the schism (materials)   & \textsc{schism} & 1,835  \\
& Missive (Ivan of Pskov)   & \textsc{pskov-ivan} & 339  \\
& Testament (Ivan Jur'evič Graznoj)   & \textsc{dux-graz} & 421  \\
& Life of Avvakum  & \textsc{avv} & 22,835  \\
& Tale of Dracula  & \textsc{drac} & 2,487  \\
& The tale of Luka Koločskij   & \textsc{luk-koloc} & 906  \\
& The taking of Pskov   & \textsc{pskov} & 2,326  \\
& The tale of the fall of Constantinople   & \textsc{const} & 9,258  \\
 & Vesti-Kuranty & \textsc{vest-kur} & 1,154  \\
 & Zadonščina & \textsc{zadon} & 2,399  \\

ONov  & Birchbark letters  & \textsc{birchbark} & 1,965  \\
& Novgorod service book marginalia   & \textsc{nov-mar} & 93  \\
& Novgorodians' losses    & \textsc{nov-list} & 187 \\
    \hline
  \end{tabular}
\end{table}

\end{document}